\documentclass[runningheads]{llncs}
\usepackage[T1]{fontenc}
\usepackage[colorlinks]{hyperref}
\usepackage{graphicx}
\usepackage{amsmath}
\usepackage{multirow}
\usepackage{longtable}
 \usepackage{amssymb}
\usepackage{lipsum}
\usepackage{xcolor}
\usepackage{subfigure}
\usepackage{graphicx}
\usepackage{pifont}
\usepackage{marvosym}

\begin{document}

\title{HySparK: Hybrid Sparse Masking for Large Scale Medical Image Pre-Training}
\titlerunning{HySparK}

\author{
Fenghe Tang\inst{1,2} 
\and Ronghao Xu\inst{1,2}  
\and Qingsong Yao\inst{3}  
\and Xueming Fu\inst{1,2}  
\and Quan Quan\inst{3}  
\and Heqin Zhu\inst{1,2}  
\and Zaiyi Liu\inst{4,5}  
\and S. Kevin Zhou\inst{1,2,3} 
$^{\href{mailto:s.kevin.zhou@gmail.com}{\textrm{\Letter}}}$} 

\authorrunning{Fenghe Tang et al.}

\institute{School of Biomedical Engineering, Division of Life Sciences and Medicine, University of Science and Technology of China, Hefei, Anhui, 230026, P.R. China\\ \and
Suzhou Institute for Advanced Research, University of Science and Technology of China, Suzhou, Jiangsu, 215123, P.R. China  \and
Key Lab of Intelligent Information Processing of Chinese Academy of Sciences
(CAS), Institute of Computing Technology, CAS, Beijing 100190, China \and 
Department of Radiology, Guangdong Provincial People’s Hospital, Guangdong Academy of Medical Sciences, Guangzhou, China \and
Guangdong Provincial Key Laboratory of Artificial Intelligence in Medical Image Analysis and Application, Guangdong Provincial People’s Hospital, Guangdong Academy of Medical Sciences, Guangzhou, China
}

\maketitle              
\begin{abstract}
The generative self-supervised learning strategy exhibits remarkable learning representational capabilities. However, there is limited attention to end-to-end pre-training methods based on a hybrid architecture of CNN and Transformer, which can learn strong local and global representations simultaneously. To address this issue, we propose a generative pre-training strategy called \textbf{Hy}brid \textbf{Spar}se mas\textbf{K}ing (HySparK) based on masked image modeling and apply it to large-scale pre-training on medical images. First, we perform a bottom-up 3D hybrid masking strategy on the encoder to keep consistency masking. Then we utilize sparse convolution for the top CNNs and encode unmasked patches for the bottom vision Transformers. Second, we employ a simple hierarchical decoder with skip-connections to achieve dense multi-scale feature reconstruction. Third, we implement our pre-training method on a collection of multiple large-scale 3D medical imaging datasets. Extensive experiments indicate that our proposed pre-training strategy demonstrates robust transfer-ability in supervised downstream tasks and sheds light on HySparK's promising prospects. The code is available at \url{https://github.com/FengheTan9/HySparK}.

\end{abstract}

\section{Introduction}

Due to the scarcity of time-consuming and labor-intensive labeled medical images, pre-training on large amounts of easy-collected unlabeled medical images by self-supervised learning approaches to learn representations for downstream tasks is a promising approach in medical image analysis (MIA)~\cite{sslmia}.
Self-supervised learning approaches can be divided into two families: Contrastive methods~\cite{simclr,moco,byol,swav,simsiam,dino} and Generative methods~\cite{bert,gpt,inpaint,beit,ibot,mae,simmim,jepa,cmae,spark}, where the latter group demonstrates better transferability to downstream tasks~\cite{mae,spark} such as segmentation. Representative generative methods like MAE~\cite{mae} pre-train the Vision Transformers (ViTs)~\cite{vit} in "BERT-style"~\cite{bert} by dropping masked non-overlapping patches and re-predicting the masked ones.

From the architecture perspective, the inductive bias of CNN~\cite{u-net} and the long-range representation ability of Transformer~\cite{vit} play pivotal roles in achieving excellent performance in visual tasks. However, the local limitation of CNNs constrains their ability to overcome performance bottlenecks further. Additionally, due to the scarcity and sparsity of medical images, the limited inductive biases and data-hungry nature of ViTs~\cite{vit} make it challenging to effectively transfer to downstream tasks~\cite{medmae,swinunetr}. To integrate the advantages of both worlds at the infrastructure design level, a hybrid architecture leverages the inductive bias of CNNs and the global context learning capabilities of ViTs, showing great potential to break-through the performance bottlenecks on medical images~\cite{transunet,transbts,uctransnet,mobileutr}.

Based on this advancement, a natural insight arises: Is it possible to simultaneously pre-train CNN and ViT with large-scale unlabeled medical images, which fully capitalize on the advantages of the hybrid model to unleash its potential? Despite MAE~\cite{mae} is able to pre-train ViTs~\cite{vit}, for CNNs, executing sliding windows can erode the masked regions, leading to a vanishing mask pattern and causing a pixel distribution shift issue~\cite{spark}. Luckily, SparK~\cite{spark} successfully extends the masked image modeling to CNNs by deploying sparse convolution~\cite{spconv} to calculate only unmasked positions and skip the masked pixels.

Nevertheless, extending the success of the "BERT-style" masked imaging modeling pre-training strategy from single to hybrid architectures remains a challenging yet unrealized problem. Two main challenges are hindering the end-to-end implementation of pre-training hybrid architectures: \textbf{(i) Masking consistency.}  For a single architecture, it is easy to maintain masking consistency~\cite{beit,ibot}\\~\cite{mae,simmim,cmae,spark}. However, for hybrid architectures, due to the inconsistent masking strategies in both worlds, the direct combination still causes the "pixel distribution shift" and "mask pattern vanishing" issues~\cite{spark}. \textbf{(ii) Multi-scale representation learning is necessary.} In medical imaging, a series of u-shape networks such as U-Net~\cite{u-net} demonstrate the importance of multi-scale and skip-connections in improving model performance. However, most current algorithms only learn representations at a single scale~\cite{mae,simmim}, neglecting the performance advantages brought by multi-scale architectures, which is crucial in MIA tasks. Although SparK~\cite{spark} takes this issue into account, it uses a simple fusion method (only skip-addition) and ignores important pattern adaptation in downstream tasks (success of skip-connections in medical downstream tasks~\cite{u-net,mednext,unetr,swinunetr}), which widen the gap from pre-training to downstream transferring.

In this work, we address the above issues and propose a \textbf{Hy}brid \textbf{Spar}se mas\textbf{K}ing (HySparK) strategy for self-supervised learning in CNN and Transformer hybrid architectures. Following the success of hybrid architectures in MIA tasks~\cite{transunet,transbts,uctransnet,mobileutr}, we use CNN as the top encoder to extract local representations and Transformer as the bottom for global features. Moreover, skip-connections are introduced to integrate multi-scale representations. To address \textbf{masking consistency}, we perform \textbf{bottom-up} masking. Specifically, we initialize the masks based on the patch division in the bottom ViT layer, which are then mapped to the upper CNN layers with different scales. Then, we use sparse convolution in CNN and drop masked patches in Transformer to avoid calculating mask regions. This novel design ensures the consistency of mask mapping between different layers in the hierarchical CNN and ViT, which prevents the data distribution shift problem~\cite{spark} in the hybrid encoder.

In the decoding stage, to leverage the advantage of skip-connections mentioned in issue \textbf{(ii)}, we construct a simple hierarchical decoder. For the skip-connections (concat and fuse), we fill the mask embeddings into all empty positions of multi-scale features. Finally, we reconstruct the masked pixels. To our knowledge, HySparK is the first successful generative-based 3D hybrid architecture method for self-supervised learning, applied to large-scale 3D CT medical image pre-training. Similar to SparK~\cite{spark}, HySparK is a general method that does not restrict the specific hybrid encoder (e.g. specific CNN or Transformer) to pre-train. In this paper, we utilize the representative CNN network in the medical imaging analysis, modern MedNeXt~\cite{mednext}, as the top CNN. For the bottom Transformer, we employ the standard ViT~\cite{vit}. Across multiple segmentation downstream tasks, HySparK outperforms state-of-the-art medical self-supervised pre-training methods and single-architecture-based masking approaches like MAE~\cite{mae}. Our primary contributions are as follows:

\begin{enumerate}
    \item 
    We propose a generative self-supervised learning method to {\bf pre-train a hybrid architecture}, which unleashes its strengths to integrate both local and global representations. We are the {\bf first} to pre-train the hybrid architecture in an end-to-end fashion.

    \item 
    We design the bottom-up {\bf 3D hybrid masking} to keep the consistency of mask modeling and data distribution across different architectures.
    
    \item 
    We pre-train a strong hybrid vision encoder using HySparK on large-scale CT medical image datasets (6.8K CT scans in total). Extensive experiments on downstream tasks demonstrate the {\bf effectiveness and potential} of HySparK in its transferability.
\end{enumerate}

\section{Approach}

\begin{figure}[t]
\includegraphics[width=\textwidth]{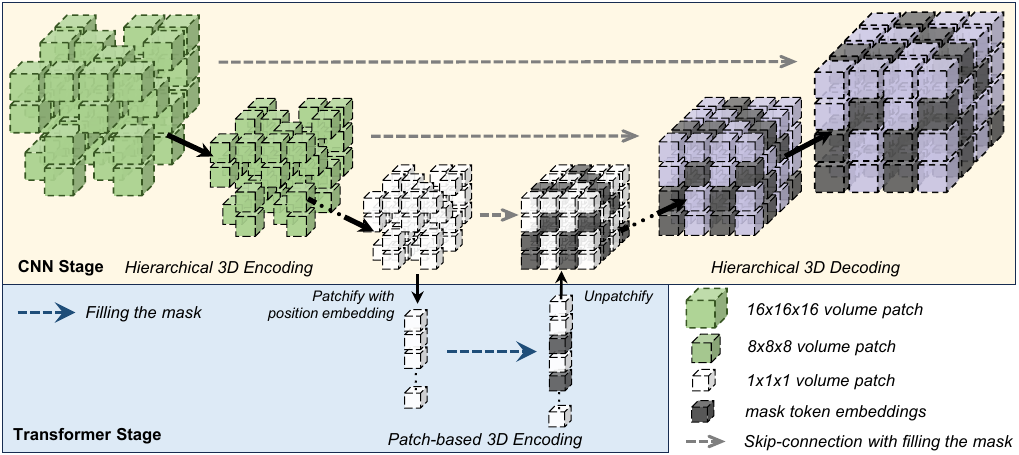}
\caption{Hybrid Sparse masKing (HySparK). The hybrid architecture comprises a CNN at the top (\textcolor{yellow}{yellow}) and a Transformer at the bottom (\textcolor{blue}{blue}). We initiate the masking strategy at the junction between the CNN and Transformer and execute bottom-up mask modeling. The initialization unmasking patch is white, the bottom-up mapping unmasking patch is \textcolor{green}{green} and the masking position is \textbf{black}.} \label{fig1}
\end{figure}

As shown in Fig.~\ref{fig1}, our proposed HySparK framework aims to mask a portion of the image through hybrid masking and pre-train the encoder by reconstructing the masked patches. The HySparK framework comprises two stages: the CNN stage and the Transformer stage. Firstly, the bottom-up sparse masking is performed in the encoder (Section~\ref{sec:sec1}), where hierarchical 3D encoding is conducted in the upper CNN stage, while patch-based 3D encoding takes place in the bottom Transformer stage. Secondly, hierarchical 3D decoding is conducted using a hierarchical decoder with skip-connections (Section~\ref{sec:sec2}) to learn multi-scale representations. Finally, we describe the pre-train optimization objectives of HySparK (Section~\ref{sec:sec3}).

\subsection{Hybrid masking}
\label{sec:sec1}
We perform hybrid masking in a bottom-up manner. Specifically, we divide the encoder into a top-level $N$-stage CNN encoder $[\mathit{E}^{cnn}_{1\sim N}]$ ({\it e.g.}, $N=4$ stages ResNet-style~\cite{resnet} or ConvNeXt-style~\cite{convnext} encoder) and a bottom-level ViT encoder $\mathit{E}^{tr}$. We initialize sparse mask $M_n \in \mathbb{R}^{\frac{H}{n^2} \times \frac{W}{n^2}\/  \times \frac{D}{n^2}}$ at the junction of the two architectures (i.e. initializing masking at the output of the last CNN layer, before ViT). To maintain consistency in masking across different architectures, we ensure that both architectures adhere to the junction-initialized masking.

\noindent\textbf{Hierarchical 3D encoding in CNN stage.} Since the masking is initialized at the last layer of the CNN, we upsample the initialized $M_n$ sparsely backward to different CNN stages, generating a set of mask $[M_{1\sim n-1}]$ at different scales with the same rules from $M_n$. Subsequently, we utilize 3D sparse convolutions to generate different scales sparse features $[S_{1\sim n}]$ with masking $[M_{1\sim n}]$ and feature maps $[f^{cnn}_{1\sim n}]$:
\begin{equation}
\label{EQ:sp}
    \text{SparseConv}(f^{cnn}_i, M_i) \to S_i, \quad  \forall i\in\{1,2,...,N\}.
\end{equation}

\noindent\textbf{Patch-based 3D encoding in Transformer stage.} As the bottom encoder is a standard ViT, learning only unmasked patches. We divide the features obtained from the last layer output of the CNN into patches with position embeddings. Subsequently, following the initialized masking rules, we remove the masked patches and only utilize the tokens $T$ without masking:
\begin{equation}
\label{EQ:tr}
    \text{Patchify}(S_n, M_n) \to T.
\end{equation}

\subsection{Hierarchical decoding with skip connections}
\label{sec:sec2}
In decoding stage, we introduce a simple cascaded decoder comprising $N-1$ upsampling blocks $\{B_{1}^{up}, B_{2}^{up},...,B_{n-1}^{up}\}$ and $N-1$ fusion blocks $\{B_{1}^{f}, B_{2}^{f},...,B_{n-1}^{f}\}$ for skip connections. Before decoding, we first unpatchify the tokens $T$ from the Transformer stage into sparse feature map $S_{n}$. Next, we fill mask embeddings into all empty positions of the sparse features at different scales to get dense features $[S_{1\sim n}^{'}]$. After applying projection layers to reduce the width of dense features at different scales, we perform hierarchical decoding with skip connection via:
\begin{equation}
\label{EQ:decode1}
    D_n  = \phi_{n}(S_{n}^{'}).
\end{equation}
\begin{equation}
\label{EQ:decode1}
    D_i  = B_{i}^{f}(\text{Concat}\{B_{i}^{up}(D_{i+1}),\phi_{i}(S_{i}^{'})\}), \quad (\forall i\in\{N-1,...,2,1\}).
\end{equation}
where $\phi_{i}$ denotes the linear projection layer of the $i$th stage. $D_n$ and $D_i$ represent the input and output of the decoder. The final output of the decoder is $D_1$.

\subsection{Optimization objectives and downstream fine-tuning}
\label{sec:sec3}
We utilize a linear layer to reconstruct $D_1$. Moreover, similar to MAE~\cite{mae} and SparK~\cite{spark}, a mean square error loss ($\mathcal{L}_2$) is used for reconstruction optimization of normalized pixels at masked positions. During fine-tuning, we only use the encoder to accomplish downstream tasks without any adjustment, as dense input is a special case of sparse input~\cite{spark}. We use a combined loss ($\mathcal{L}_{seg}$) of binary cross entropy ($\mathcal{L}_{BCE}$) and Dice loss ($\mathcal{L}_{Dice}$) to optimize the network.

\section{Experiment}

\subsection{Datasets}
\noindent\textbf{Pre-training datasets}: 
A total of 13 public CT datasets, consisting of \textbf{6,814} CT scans, are curated to form our pre-training dataset (reviewed in Table~\ref{tab:dataset}). Existing annotations or labels are not utilized from these datasets during pre-training. The pre-train datasets are interpolated to the isotropic voxel spacing of $1.5\ mm$. Intensities are scaled to $[-175, 250]$, then normalized to $[0, 1]$. We crop sub-volumes of $96\times96\times96$ voxels.

\noindent\textbf{BTCV dataset}: The BTCV dataset~\cite{btcv} consists of 30 subjects with abdominal CT scans where 13 organs are annotated by interpreters under supervision of clinical radiologists at Vanderbilt University Medical Center. Our data preprocessing strategy is the same as UNETR~\cite{unetr}.

\noindent\textbf{MSD datasets}: Medical Segmentation Decathlon (MSD) dataset~\cite{msd} comprises ten segmentation tasks from different organs and image modalities. We only use six CT datasets: Liver, Lung, Pancreas, Hepatic Vessel, Spleen, and Colon datasets. All the pre-processing strategies are the same as Swin UNETR~\cite{swinunetr}.

\subsection{Settings}
HySparK can use any 3D convolutional network and patch-based ViT as the hybrid architecture's encoder. In the CNN stage, we choose MedNeXt~\cite{mednext} (the state-of-the-art ConvNet in medical tasks) as the top encoder. In the Transformer stage, we implement the standard ViT~\cite{vit} as the bottom encoder. It is worth noting that we substitute the downsampling layers of MedNeXt with max-pooling. Additionally, for the pre-trained decoder, the upsampling block consists of two convolutional layers and an upsampling layer, while the fusion block comprises two convolutional layers. For downstream tasks, we utilize the MedNeXt decoder for segmentation.

For pre-training tasks, we train with an AdamW optimizer, an initial learning rate of 1e-4, and a cosine-annealing learning rate scheduler. The pre-training experiments use a batch-size of 8 on a single GPU and 100 epochs in 4 days. For downstream segmentation tasks, a five-fold cross-validation strategy is used to train models for all BTCV and MSD experiments and we select the best model in each fold. Detailed training hyperparameters for fine-tuning BTCV and MSD tasks are the same as Swin UNETR~\cite{swinunetr}. All methods are implemented in PyTorch and trained on an Nvidia A800.

The Dice similarity coefficient (Dice) is used as the measurement for experiment results. We select three advanced generative-based self-supervised learning strategies: Transformer-based MAE~\cite{mae} and SimMIM~\cite{mae}, CNN-based SparK~\cite{spark} and two advanced contrastive-based self-supervised learning method: Swin UNETR~\cite{swinunetr} Pre-trained method (SUP) and vox2vec~\cite{vox2vec}. In addition, we choose the current well-known segmentation networks UNETR~\cite{unetr}, Swin UNETR~\cite{swinunetr}, and MedNeXt~\cite{mednext} as the downstream segmentation task networks of MAE, SimMIM, and SparK, respectively. It is worth noting that MAE, SimMIM, and SparK methods are obtained by using official codes and extending them to 3D.

\begin{table}[!t]
\caption{Overview of Pre-train Dataset.\label{tab:dataset}}
\resizebox{1\linewidth}{!}
{
\begin{tabular}{l rr c | l rr c }
\hline 
Dataset (year) & \# of classes & \# of volumes & downstream & Dataset (year) & \# of classes & \# of volumes & downstream \\
\hline
BTCV (2015)~\cite{btcv} & 13 & 50 &  \checkmark &  MSD Liver (2021)~\cite{msd} & 2 & 201 & \checkmark\\
CHAOS (2018)~\cite{chaos} & 4 & 40 & & MSD Lung (2021)~\cite{msd} & 2 & 95 & \checkmark \\
WORD (2021)~\cite{word} & 16 & 150 &  & MSD Pancreas (2021)~\cite{msd} & 2 & 420 & \checkmark\\
FLARE'22 (2022)~\cite{flare22} & 13 & 2300 & &  MSD Hepatic Vessel (2021)~\cite{msd} & 1 & 443 & \checkmark \\
AbdomenCT-1k  (2022)~\cite{abd1k} & 4 & 1062 & & MSD Spleen (2021)~\cite{msd} & 1 & 61 & \checkmark \\
TotalSegmentator (2022)~\cite{totalseg} & 104 & 1202  & &  MSD Colon (2021)~\cite{msd} & 1 & 190 & \checkmark \\
\cline{5-8}
AMOS22 (2022)~\cite{amos} & 15 & 600 &  & \multicolumn{1}{l}{Total} & & 6814\\

\hline
\end{tabular}
}
\end{table}

\subsection{Results and discussion}


\noindent\textbf{Results on BTCV dataset.} Evaluation results on BTCV are shown in Table~\ref{tab:btcv}. Compared with other competitive methods, the proposed HySparK achieves the best performance. We obtain the highest average Dice of 80.67\%, which at least improves by 1.17\% compared to other baselines. Additionally, we achieve significant improvements in segmenting organs with smaller sizes, such as the pancreas and adrenal glands. This shows that HySparK effectively learns strong multi-scale representations. In addition, we fine-tune the pre-trained models using a smaller (20\%) training set, our HySparK significantly outperforms state-of-the-art methods (average Dice of 64.27, 2.5\% higher than other methods) and gains the best trade-off performance in different scale organs, which highlights the powerful downstream transferring capability of our method.

\begin{table}[!t]
\caption{Result on BTCV. \textbf{val} (bold) / \underline{val} (underline) : top method / second method.\label{tab:btcv}}
\resizebox{1\linewidth}{!}
{
\begin{tabular}{l | l | l | ccccccccccc| c}
\hline 
\multirow{2}{*}{} & \multicolumn{2}{c |}{Pre-training Method}  & \multirow{2}{*}{Spl} & \multirow{2}{*}{Kid} & \multirow{2}{*}{Gall} & \multirow{2}{*}{Eso} & \multirow{2}{*}{Liv} & \multirow{2}{*}{Sto} & \multirow{2}{*}{Aor} & \multirow{2}{*}{IVC} & \multirow{2}{*}{Veins} & \multirow{2}{*}{Pan} & \multirow{2}{*}{AG} & \multirow{2}{*}{Avg} \\
\cline{2-3}
& \multicolumn{1}{c| }{ Method} & \multicolumn{1}{c| }{Network} &&&&&&&&&&&& \\
\hline


\multirow{6}{*}{20\%} &  vox2vec~\cite{vox2vec} & 3D UNet(FPN)~\cite{u-net} & 73.97 & 66.80 & 34.64 & 49.51 & 86.72 & 54.28 & 73.70 & 64.43 & 41.64 & 37.35 & 27.00 & 54.14 \\

& SUP~\cite{swinunetr} & Swin UNETR~\cite{swinunetr} & 71.29 & 59.78 & 40.43 & 57.30 & 87.91 & 51.24 & 70.88 & 59.13 & 50.16 & 37.70 & 34.24  & 54.93 \\


& MAE~\cite{mae} & UNETR~\cite{unetr} & 66.22 & \textbf{78.63} & \textbf{48.19} & 24.02 & \textbf{92.51} & \textbf{78.50} & 79.00 & \textbf{78.28} & 35.63 & \underline{52.99} & 18.52 & 57.91 \\


& SimMIM~\cite{mae}   & Swin UNETR~\cite{swinunetr} & 71.16 & 62.76 & 42.32 & 56.05 & 88.45 & 52.46 & 71.95 & 60.94 & 51.32 & 39.39 & \underline{35.10} & 56.14\\


& SparK~\cite{spark} & MedNeXt~\cite{mednext} & \underline{79.78} & \underline{72.29} & 38.66 & \underline{58.89} & \underline{91.71} & \underline{68.68} & \textbf{81.24} & \underline{71.93} & \underline{57.13} & 51.96 & 29.01 & \underline{61.74} \\


& HySparK & MedNeXt+ViT & \textbf{79.84} & 70.54 & \underline{46.03} & \textbf{60.86} & 91.29 & 67.12 & \underline{79.42} & 69.21  & \textbf{59.07} & \textbf{54.34} & \textbf{43.61}  & \textbf{64.27} \\

\hline

\multirow{6}{*}{100\%} &  vox2vec~\cite{vox2vec} & 3D UNet(FPN)~\cite{u-net} & \textbf{91.40} & \textbf{90.70} & 59.50 & 72.70 & \textbf{96.30} & 83.20 & \textbf{91.30} & \underline{83.90} & 69.20 & 73.90 & \underline{65.20} & \underline{79.50} \\
& SUP~\cite{swinunetr} & Swin UNETR~\cite{swinunetr} & 84.20 & 86.70 & 58.40 & 70.40 & 94.50 & 76.00 & 87.70 & 82.10 & 67.00 & 69.80 & 61.00 & 75.80 \\
& MAE~\cite{mae} & UNETR~\cite{unetr} & 90.71 & 87.63 & \underline{62.50} & 70.69 & 94.73 & \underline{86.11} & 90.59 & 83.26 & \textbf{71.00} & 75.47 & 63.77 & 79.07 \\
& SimMIM~\cite{mae}   & Swin UNETR~\cite{swinunetr} & 87.12 & 80.85 & 60.28 & 72.34 & 93.70 & 78.42 & 87.89 & 81.46 & 64.92 & 66.34 & 58.65 & 74.73\\
& SparK~\cite{spark} & MedNeXt~\cite{mednext} & 90.02 & 87.78 & 62.48 & \textbf{74.36} & 95.00 & 84.85 & 90.17 & 83.60 & 68.83 & \underline{76.57} & 64.13 & 79.21 \\
& HySparK & MedNeXt+ViT & \underline{90.67} & \underline{88.32} & \textbf{68.18} & \underline{74.20} & \underline{95.03} & \textbf{87.46} & \underline{90.17} & \textbf{84.50} & \underline{70.04} & \textbf{78.36} & \textbf{66.75} & \textbf{80.67} \\

\hline
\end{tabular}
}

\end{table}

\begin{table}[!t]
\caption{Result on MSD. \textbf{val} (bold) / \underline{val} (underline) : top method / second method.\label{tab:msd}}
\resizebox{1\linewidth}{!}
{
\begin{tabular}{l | l |rrr | r | rrr | rrr | r | r | c}
\hline 

\multicolumn{2}{c}{Pre-training method} 
& \multicolumn{3}{|c}{ Liver}
& \multicolumn{1}{|c}{ Lung}
& \multicolumn{3}{|c}{ Pancreas}
& \multicolumn{3}{|c}{Hepatic Vessel}
& \multicolumn{1}{|c}{ Spleen}
& \multicolumn{1}{|c|}{ Colon}
&  \multirow{2}{*}{Avg}
\\
\cline{1-14}
\multicolumn{1}{c|}{Method} & \multicolumn{1}{c|}{Network} & \multicolumn{1}{c}{Dice1} & \multicolumn{1}{c}{Dice2}  & \multicolumn{1}{c|}{Avg}  & \multicolumn{1}{c|}{Dice} & \multicolumn{1}{c}{Dice1} & \multicolumn{1}{c}{Dice2} & \multicolumn{1}{c|}{Avg} & \multicolumn{1}{c}{Dice1} & \multicolumn{1}{c}{Dice2} & \multicolumn{1}{c|}{Avg} & \multicolumn{1}{c|}{Dice} & \multicolumn{1}{c|}{Dice} &\\
\hline
vox2vec~\cite{vox2vec}             & 3D UNet(FPN)~\cite{u-net}              & 95.60 & 51.00 & 73.70 & 56.60 & 77.00 & 31.80 & 54.40 & 59.50 & 62.40 & 60.95 & 96.10 & 30.10  & 61.97\\
SUP~\cite{unetr} & Swin UNETR~\cite{unetr} & 95.00 & 49.30 & 72.15 & 55.20 & 75.20 & 35.90 & 55.55 & 60.90 & 57.50 & 59.20 & 95.50 & 29.20 & 61.13 \\
MAE~\cite{mae}      & UNETR~\cite{unetr}         & 95.49 & 56.47 & 75.98 & 56.42 & 77.76 & 39.29 & 58.52  & 59.99 & 62.22 & 61.10  & 95.28 & 34.53 & 63.63 \\
SimMIM~\cite{mae}   & Swin UNETR~\cite{unetr}    & 95.32 & 55.25 & 75.28 & 60.31 & 76.16 & 44.96 & 60.56 & 60.67 & 61.79 & 61.23 & 95.64 & 41.11 & 65.68 \\
SparK~\cite{spark}  & MedNeXt~\cite{mednext}     & \underline{95.87} & \textbf{62.95} & \textbf{79.41} & \underline{65.58} & \underline{78.88} & \underline{47.86} & \underline{63.37}  & \underline{61.08} & \underline{67.76} & \underline{64.42} & \underline{96.18} & \underline{49.85} & \underline{69.80} \\
HySparK             & MedNeXt+ViT                & \textbf{96.02} & \underline{60.92} & \underline{78.47} & \textbf{65.96} & \textbf{79.69} & \textbf{49.67} & \textbf{64.68} & \textbf{61.58} & \textbf{69.36} & \textbf{65.47} & \textbf{96.39} & \textbf{50.78} & \textbf{70.29}  \\
\hline
\end{tabular}
}
\end{table}

\begin{figure}[!t]
\centering
\begin{minipage}[t]{0.32\textwidth}
\centering
\includegraphics[width=4cm]{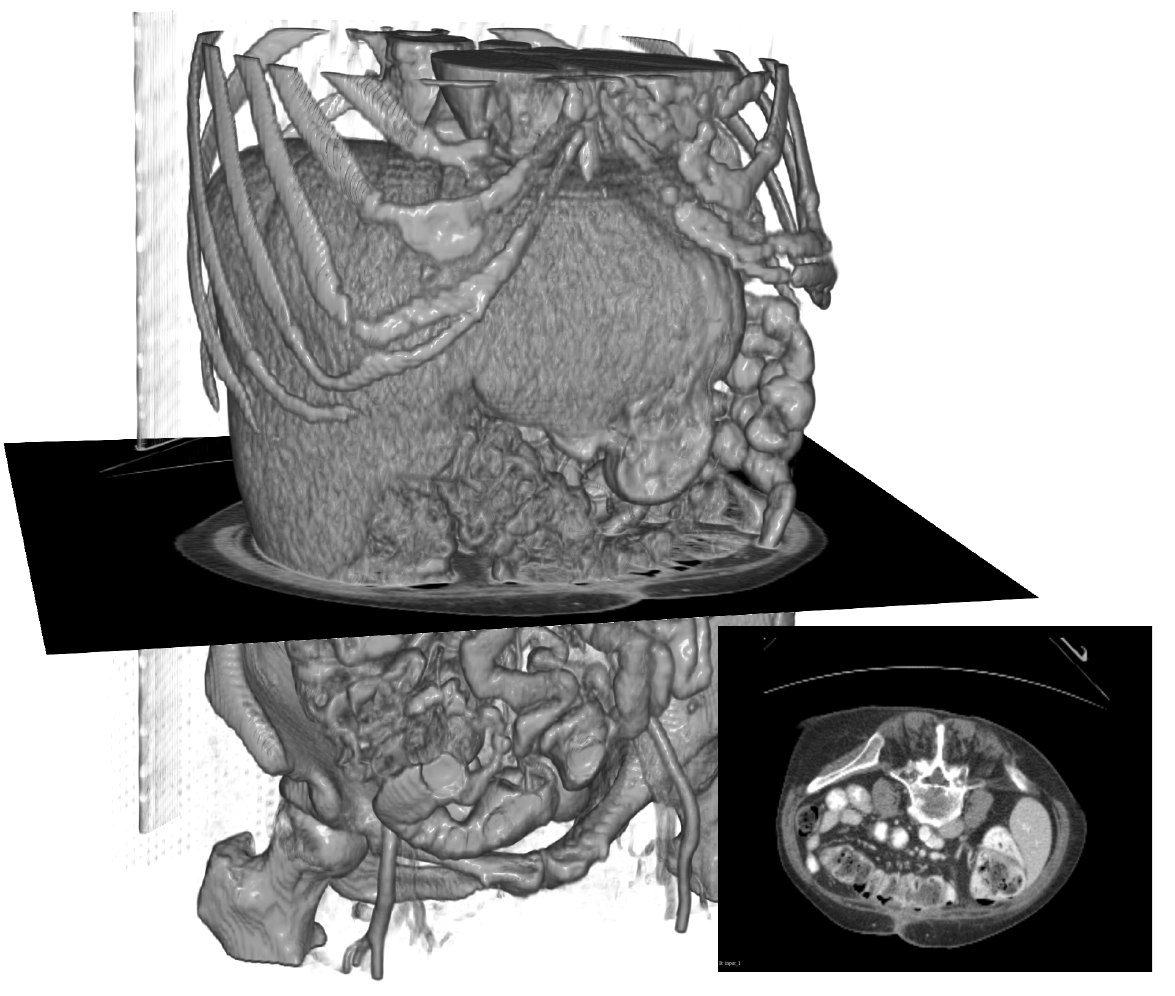}
\textbf{(a)} Raw Input
\end{minipage}
\begin{minipage}[t]{0.32\textwidth}
\centering
\includegraphics[width=4cm]{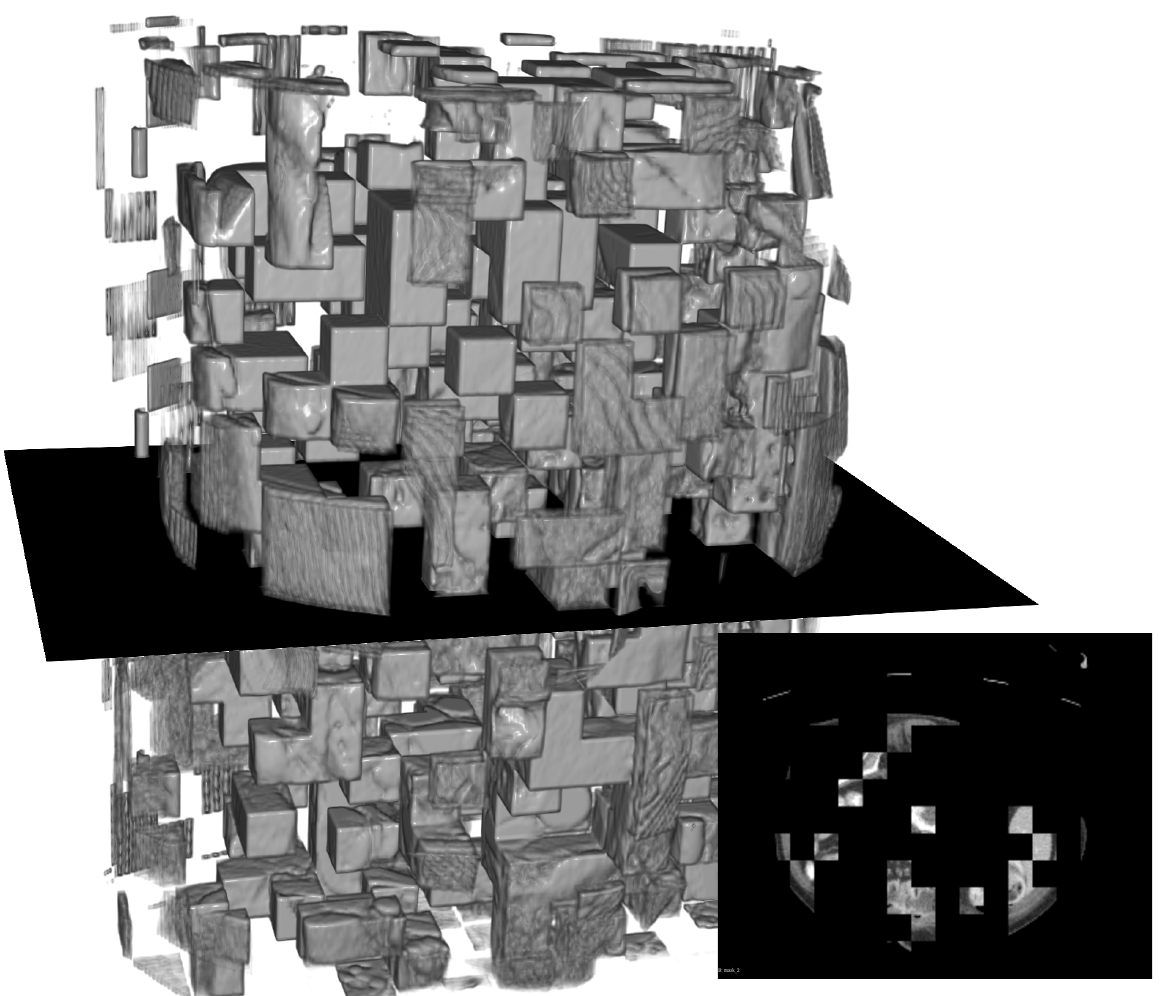}
\textbf{(b)} Masked Input
\end{minipage}
\begin{minipage}[t]{0.32\textwidth}
\centering
\includegraphics[width=4cm]{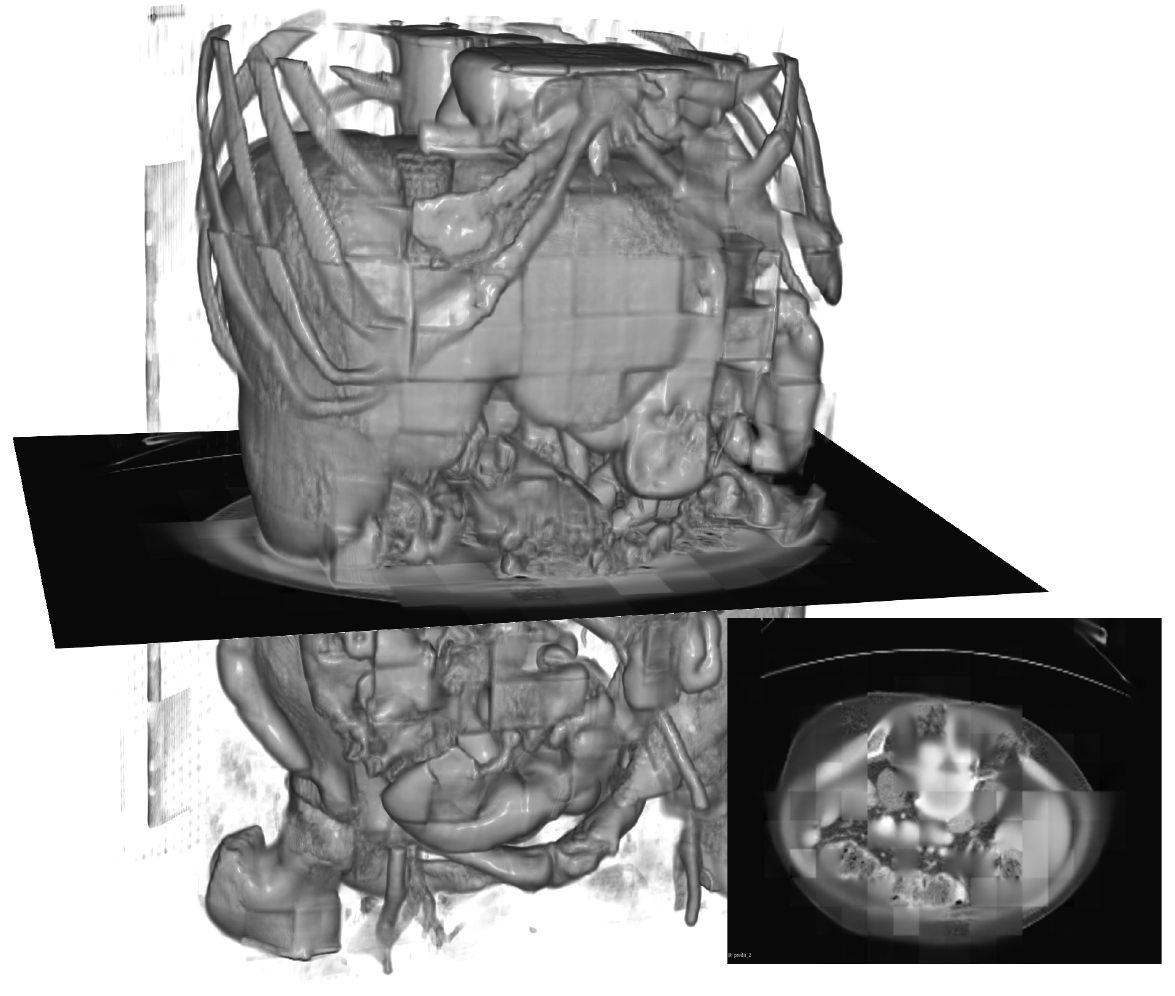}
\textbf{(b)} Prediction
\end{minipage}
\caption{Reconstruction Result by HySparK. \label{fig:re}}
\end{figure}

\noindent\textbf{Results on MSD datasets.} The overall results on the MSD dataset per task are shown in Table~\ref{tab:msd}. HySparK presents the best average Dice of 70.29\%. Our method outperforms other SOTA approaches in Lung, Pancreas, Hepatic Vessel, Spleen, and Colon tasks. Moreover, HySparK improves Pancreas and Hepatic Vessel lesions by at least 1.81\% and 1.60\% which is attributed to its strong multi-scale representation. It is worth noting that almost all generative-based methods outperform contrast-based methods, indicating the superior transferability of generative-based methods.

\noindent\textbf{Visualization.}
We visualize 3D reconstruction results to check what HySparK learns in pre-training. As shown in Fig.~\ref{fig:re}, our method can almost reconstruct the different shapes of organs, bones, and other details from the very small portion of unmasked patches.

\begin{table}[t]
\caption{Ablation study on Mask Ratio.\label{tab:ablation1}}
\resizebox{1\linewidth}{!}
{
\begin{tabular}{l | rrrrrrrrrrrrr | c}
\hline 
\multicolumn{1}{l |}{Mask Ratio}  & \multirow{1}{*}{Spl} & \multirow{1}{*}{RKid} & \multirow{1}{*}{Lkid} & \multirow{1}{*}{Gall} & \multirow{1}{*}{Eso} & \multirow{1}{*}{Liv} & \multirow{1}{*}{Sto} & \multirow{1}{*}{Aor} & \multirow{1}{*}{IVC} & \multirow{1}{*}{Veins} & \multirow{1}{*}{Pan} & \multirow{1}{*}{Rad} & \multirow{1}{*}{Lad} & \multirow{1}{*}{BTCV Avg} \\
\hline
w/o pre-trained & 89.71 & 88.17 & 86.69 & 62.73 & 73.14 & 94.44 & 83.96 & 88.94 & 82.51 & 70.02 & 72.47 & 64.73 & 63.99 & 78.58\\
mask 25 \% & 90.52 & 89.76 & 87.55 & 66.42 & 74.93 & 94.98 & 86.21 & 90.61 & 84.28 & 70.96 & 77.92 & 66.26 & 66.90 & 80.56 \\
mask 50 \% & 90.76 & 88.46 & 86.33 & 65.38 & 75.50 & 95.21 & 85.43 & 90.75 & 83.57 & 71.75 & 77.47 & 66.09 & 68.02 & 80.36 \\
mask 75 \% & 90.67 & 89.35 & 87.30 & 68.18 & 74.20 & 95.03 & 87.46 & 90.17 & 84.50 & 70.04 & 78.36 & 66.46 & 67.04 & 80.67 \\
\hline
\end{tabular}
}

\end{table}

\begin{table}[!t]
\caption{Ablation study on each components in HySparK.\label{tab:ablation}}
\centering

\begin{tabular}{l | c c c | c}
\hline 
\multicolumn{1}{l |}{HySparK components}  & \multirow{1}{*}{hybrid masking} & \multirow{1}{*}{skip-connection}  & \multirow{1}{*}{skip-addition}  & BTCV Avg\\
\hline
w/o pre-trained &  — & — & — & 78.58 \\
w/o bottom-up masking & \ding{55} & \checkmark & \ding{55} & 79.47\\
w/o skip &  \checkmark & \ding{55}  & \ding{55} & 78.80 \\
w/ skip-addition &  \checkmark & \ding{55} & \checkmark &  79.97 \\
HySparK & \checkmark & \checkmark & \ding{55} & 80.67\\

\hline
\end{tabular}

\end{table}

\subsection{Ablation study}
\noindent\textbf{Ablation Study on Mask Ratio.} Table~\ref{tab:ablation1} shows the influence of different mask ratios on the model. Surprisingly, similar to MAE~\cite{mae}, it can be found that a 75\% mask ratio achieves the best performance in downstream tasks. 

\noindent\textbf{Ablation study on HySparK components.} We first remove the two most important designs in HySparK: bottom-up hybrid masking and skip connections. When mask consistency is not maintained, we observe a significant performance degradation in row 2 of Table~\ref{tab:ablation} that almost reaches the vanilla model (row 1). It suggests that \textit{inconsistency masking will lead to ineffective pre-training}. We then remove the skip design (row 3) or only use skip-addition (row 4), the performance drops significantly compared to using skip-connections (row 5), which illustrates the importance of pattern alignment for pre-training and fine-tuning tasks.

\noindent\textbf{Ablation study on architecture.} As demonstrated in Table~\ref{tab:btcv} and~\ref{tab:msd}, when the hybrid architecture drops to single architecture (i.e., CNN in SparK or ViT in MAE), their performance experiences a certain decrease compared to the hybrid architecture. This demonstrates the significant role of the hybrid architecture and its masking strategy in medical image tasks.

\section{Conclusion}
The success of hybrid architectures in medical tasks prompts us to explore their potential in downstream tasks after being well pre-trained using large-scale unlabeled medical images. In this paper, we introduce HySparK, a generative self-supervised approach to pre-training hybrid architectures, which creates a bottom-up masking modeling strategy to solve the masking inconsistency. For the problem of data distribution shift, we use sparse convolution for encoding in the CNN stage and predict the masked tokens using unmasked patches in the Transformer stage. Moreover, we introduce skip-connections to achieve pre-training and downstream task pattern alignment. HySparK brings significant performance leaps in downstream tasks and we hope our findings can inspire more work to maximize the potential of hybrid architectures in medical tasks.

\begin{credits}
\subsubsection{\ackname} Supported by Natural Science Foundation of China under Grant 62271465, Suzhou Basic Research Program under Grant SYG202338, and Open Fund Project of Guangdong Academy of Medical Sciences, China (No. YKY-KF202206).

\subsubsection{\discintname}
The authors have no competing interests to declare
that are relevant to the content of this article.
\end{credits}


%
{
    \newpage
    \centering
    \Large
    \textbf{\title\\
    \vspace{8mm} \\
    HySparK: Hybrid Sparse Masking for Large Scale Medical Image Pre-Training
    \vspace{4mm} \\
     Supplementary Material \\
    \vspace{2.0em}
    }
}

\vspace{-8mm}
\begin{figure}[!]
    \centering
    \includegraphics[width=\textwidth]{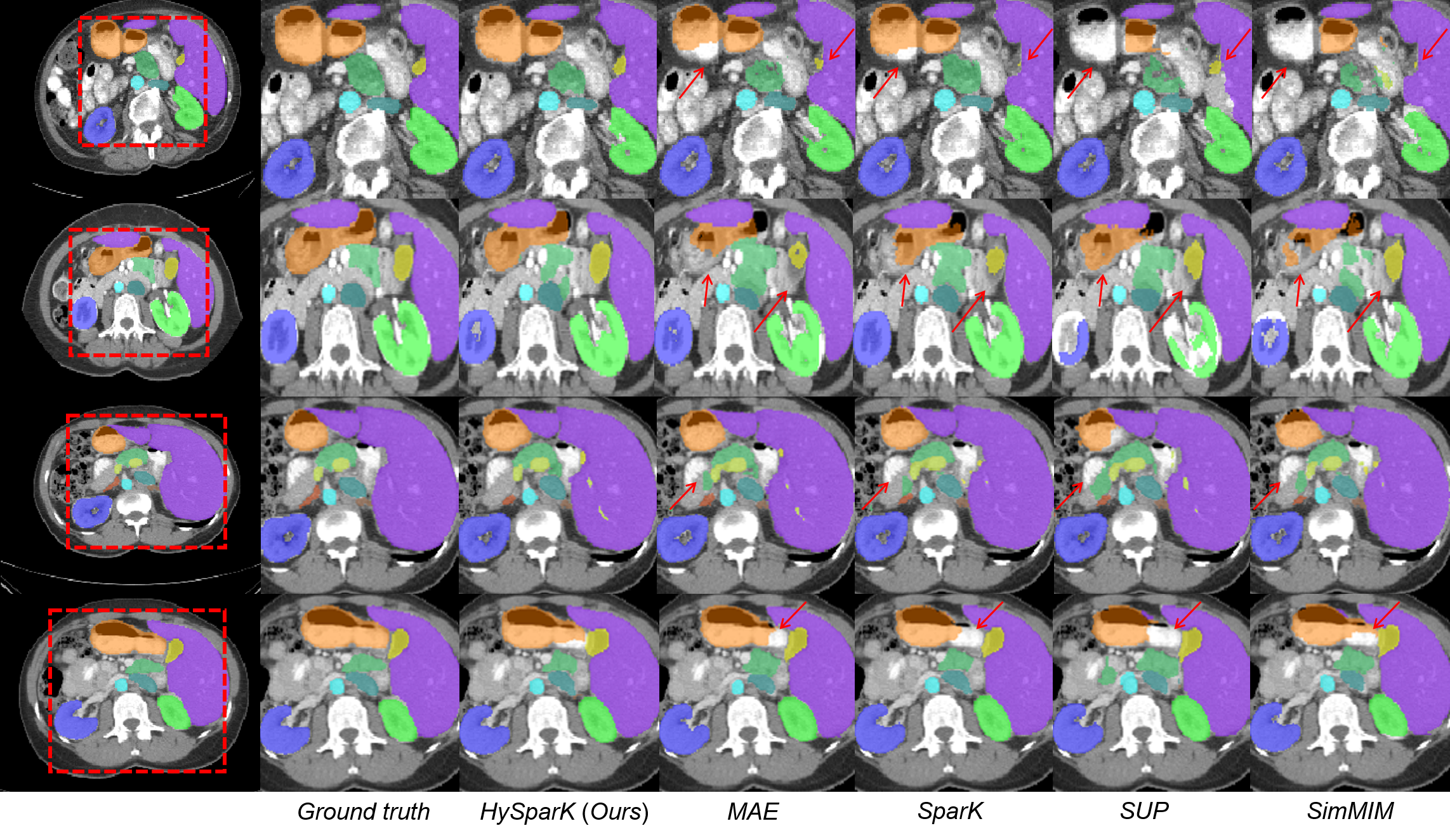}
    \vspace{-4mm}
    \caption{Visualization Results on BTCV dataset.}
    \vspace{-4mm}
    \label{fig:volume}

\end{figure}
\vspace{-2mm}
\begin{figure}[h]
\centering
\begin{minipage}[t]{0.16\textwidth}
\centering
\includegraphics[width=2cm]{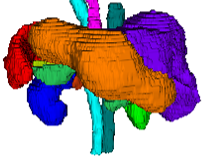}
\includegraphics[width=2cm]{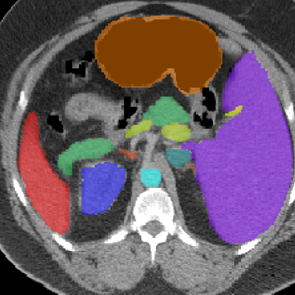}
\textbf{(a)}
\end{minipage}
\begin{minipage}[t]{0.16\textwidth}
\centering
\includegraphics[width=2cm]{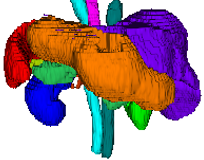}
\includegraphics[width=2cm]{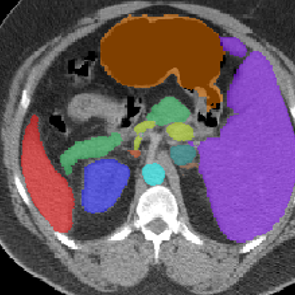}
\textbf{(b)}
\end{minipage}
\begin{minipage}[t]{0.16\textwidth}
\centering
\includegraphics[width=2cm]{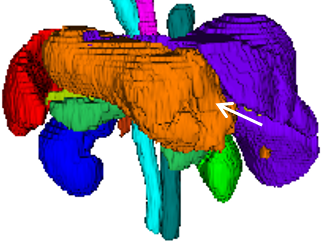}
\includegraphics[width=2cm]{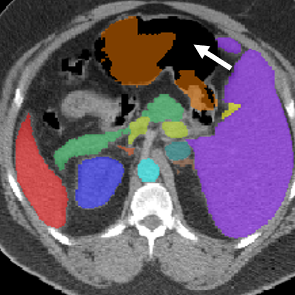}
\textbf{(c)}
\end{minipage}
\begin{minipage}[t]{0.16\textwidth}
\centering
\includegraphics[width=2cm]{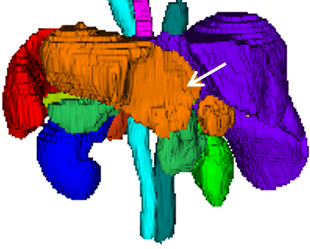}
\includegraphics[width=2cm]{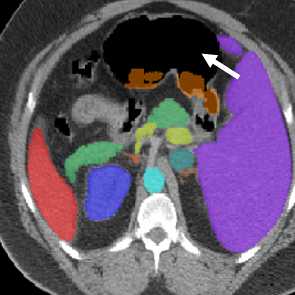}
\textbf{(d)}
\end{minipage}
\begin{minipage}[t]{0.16\textwidth}
\centering
\includegraphics[width=2cm]{3d/sp.png}
\includegraphics[width=2cm]{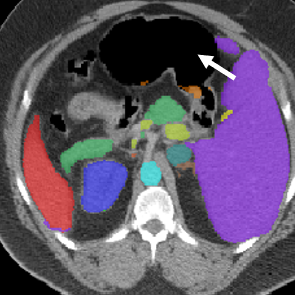}
\textbf{(e)}
\end{minipage}
\begin{minipage}[t]{0.16\textwidth}
\centering
\includegraphics[width=2cm]{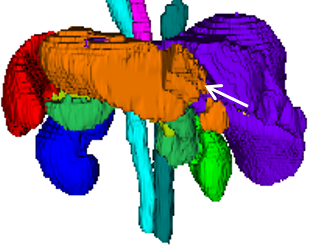}
\includegraphics[width=2cm]{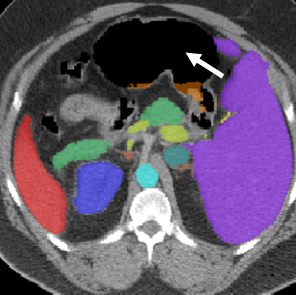}
\textbf{(f)}
\end{minipage}

\vspace{-2mm}
\caption{Visualization Results on BTCV with 3D volume. (a) Ground Truth (b) HySparK (Ours) (c) MAE (d) SparK (e) SUP (f) SimMIM. \label{fig:re}}
\end{figure}

\begin{figure}[!t]
    \vspace{-1mm}
    \centering
    \includegraphics[width=0.8\textwidth]{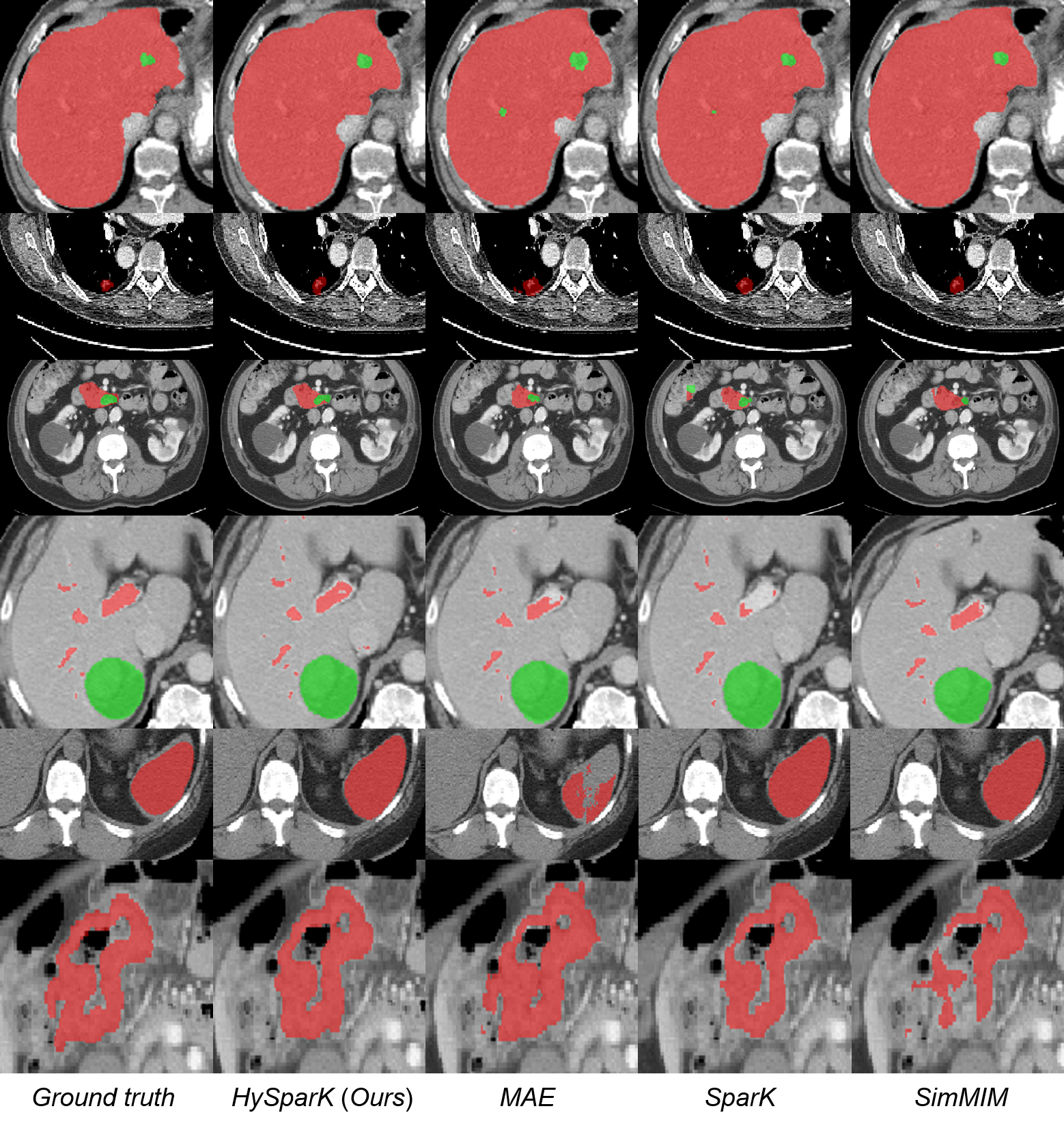}
    \vspace{-4mm}
    \caption{Visualization Results on MSD datasets. Row1 - Liver, Row2 - Lung, Row3 - Pancreas, Row4 - Hepatic Vessel, Row5 - Spleen, Row6 - Colon.}
    \label{fig:volume}

\end{figure}

\begin{figure}[h]

\vspace{-2mm}
\centering
\begin{minipage}[t]{0.16\textwidth}
\centering
\includegraphics[width=2cm]{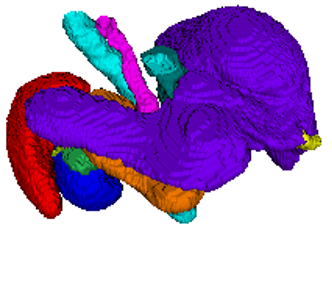}
\textbf{(a)}
\end{minipage}
\begin{minipage}[t]{0.16\textwidth}
\centering
\includegraphics[width=2cm]{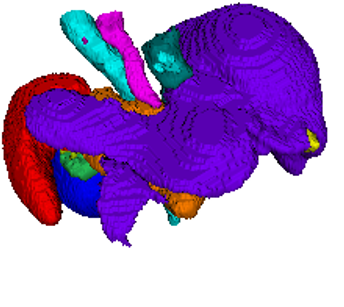}
\textbf{(b)}
\end{minipage}
\begin{minipage}[t]{0.16\textwidth}
\centering
\includegraphics[width=2cm]{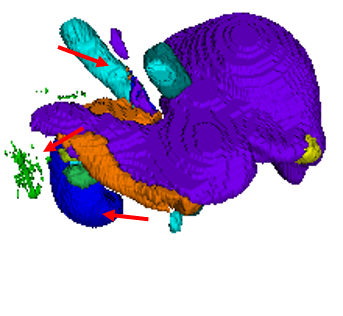}
\textbf{(c)}
\end{minipage}
\begin{minipage}[t]{0.16\textwidth}
\centering
\includegraphics[width=2cm]{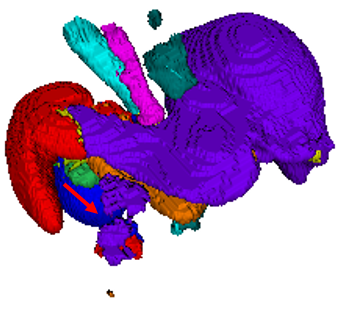}
\textbf{(d)}
\end{minipage}
\begin{minipage}[t]{0.16\textwidth}
\centering
\includegraphics[width=2cm]{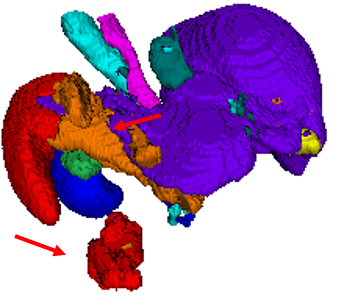}
\textbf{(e)}
\end{minipage}
\begin{minipage}[t]{0.16\textwidth}
\centering
\includegraphics[width=2cm]{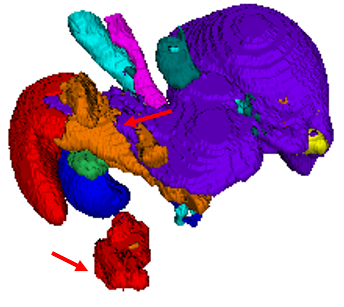}
\textbf{(f)}
\end{minipage}

\vspace{-2mm}
\caption{Visualization Results on Smaller (20\%) BTCV training set. (a) Ground Truth (b) HySparK (Ours) (c) MAE (d) SparK (e) SUP (f) SimMIM. \label{fig:re}}
\end{figure}

\begin{figure}[!]
    \vspace{-2mm}
    \centering
    \includegraphics[width=\textwidth]{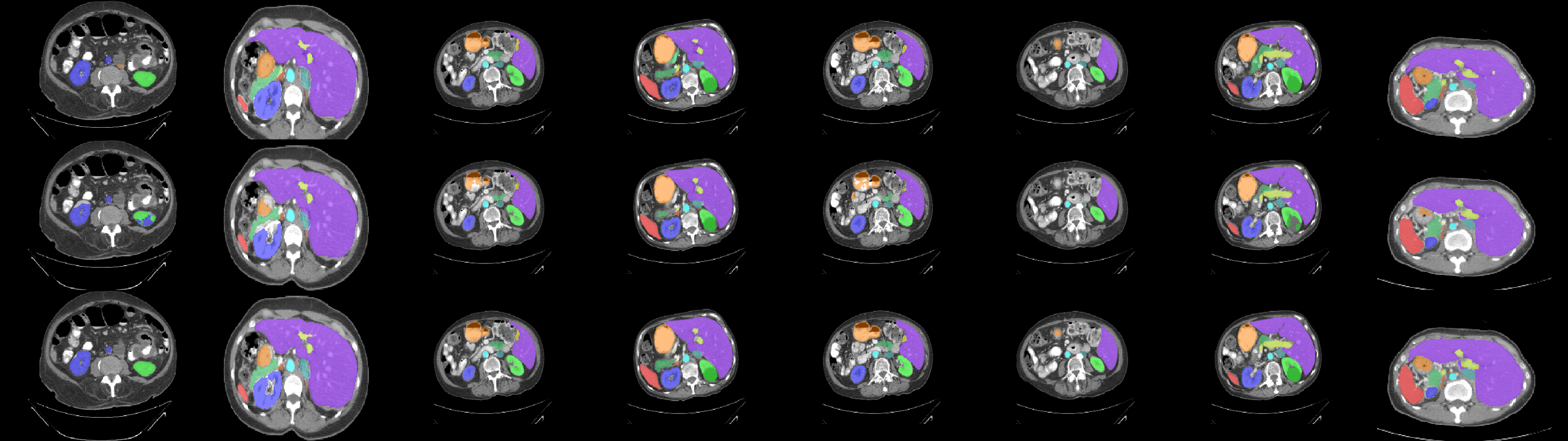}
    \vspace{-7mm}
    \caption{Visualization $w/o$ pretrain and $w/$ pretrain Results on BTCV dataset. Row1 - Ground Truth, Row2 - $w/o$ pretrain, Row3 - $w/$ pretrain.}
    \label{fig:volume}

\end{figure}

\end{document}